\documentclass{article}

\usepackage{arxiv}

\usepackage[utf8]{inputenc} 
\usepackage[T1]{fontenc}    
\usepackage{hyperref}       
\usepackage{url}            
\usepackage{booktabs}       
\usepackage{amsfonts}       
\usepackage{nicefrac}       
\usepackage{microtype}      
\usepackage{lipsum}		
\usepackage{graphicx}
\usepackage{natbib}
\usepackage{doi}
\usepackage{listings}  
\usepackage{float}  
\usepackage{xcolor} 
\usepackage{amsmath,amssymb,amsfonts}%

\usepackage{algorithm}
\usepackage{algpseudocode}
\usepackage{tabularx}
\usepackage{array}

\usepackage{multirow}%
\usepackage{rotating}

\lstdefinelanguage{yaml}{
  morekeywords={true,false,null,y,n},
  sensitive=false,
  morecomment=[l]{\#},
  morestring=[b]"
}

\definecolor{codebg}{rgb}{0.97, 0.97, 0.97}  

\title{EGRNet: A Lightweight Semantic Segmentation Network with Edge-Gated Refinement and Adversarial Sensing}

\author{
    \href{https://orcid.org/0000-0000-0000-0000}{\includegraphics[scale=0.06]{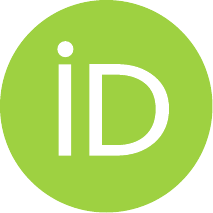}\hspace{1mm}Bareera Qaseem} \\
    School of Electrical Engineering and Computer Science (SEECS) \\
    National University of Sciences and Technology (NUST) \\
    Islamabad, Pakistan \\
    \texttt{bqaseem.msee22seecs@seecs.edu.pk} \\
    \And
    \href{https://orcid.org/0000-0002-3893-4330}{\includegraphics[scale=0.06]{orcid.pdf}\hspace{1mm}Mohsin Kamal} \\
    School of Computing \\
    University of Nebraska-Lincoln \\
    Lincoln, 68508, NE, USA \\
    \texttt{mkamal3@unl.edu} \\
    \And
    \href{https://orcid.org/0000-0002-4629-7589}{\includegraphics[scale=0.06]{orcid.pdf}\hspace{1mm}Muhammad Naveed Aman} \\
    School of Computing \\
    University of Nebraska-Lincoln \\
    Lincoln, 68508, NE, USA \\
    \texttt{naveed.aman@unl.edu} \\
}

\renewcommand{\shorttitle}{\textit{arXiv} Template}

\hypersetup{
pdftitle={A template for the arxiv style},
pdfsubject={q-bio.NC, q-bio.QM},
pdfauthor={David S.~Hippocampus, Elias D.~Striatum},
pdfkeywords={First keyword, Second keyword, More},
}

\begin{document}
\maketitle

\begin{abstract}
    As autonomous systems and smart cities continue to evolve, the demand for efficient and robust scene understanding becomes increasingly critical. Semantic segmentation plays a key role in enabling autonomous vehicles to comprehend complex urban environments. However, achieving high accuracy with minimal computational cost remains a significant challenge. In this paper, we present Edge-Gated Refinement Network (EGRNet), a lightweight and efficient deep learning model designed for real-time semantic segmentation in urban scenarios. The model incorporates depthwise separable convolutions to reduce computational complexity and dilated residual blocks for capturing rich multi-scale contextual information. Additionally, we introduce a novel Edge-Gated Refinement (EGR) module, which adaptively fuses original and refined features through a learnable gating mechanism, enhancing boundary preservation and edge-sensitive regions. To further improve feature representation, Squeeze-and-Excitation (SE) attention is applied across the network. With only 0.46M parameters, EGRNet achieves state-of-the-art performance while maintaining low computational overhead. When evaluated on the Cityscapes dataset, the model attains a mean Intersection over Union (mIoU) of 65.28\%, demonstrating strong accuracy with minimal resource consumption. Moreover, we introduce a lightweight adversarial attack detection strategy, ensuring robustness against adversarial inputs without compromising real-time performance. By combining efficiency, accuracy, and resilience, EGRNet is well-suited for deployment on edge devices in safety-critical real-time applications.
\end{abstract}


\section{Introduction}
Semantic segmentation refers to the process of assigning a semantic label to each pixel in an image. It enables fine-grained scene understanding, which is essential for numerous computer vision applications. One of the most critical use cases is autonomous driving, where accurate, real-time interpretation of road scenes is necessary for informed decision-making. Semantic segmentation facilitates safe navigation and contextual awareness by identifying key elements in urban environments, such as road markings, pedestrians, vehicles, and traffic signs.

Since the advent of end-to-end trainable architectures like SegNet, semantic segmentation has rapidly advanced \cite{guimaraes2025comparison}. Deep learning models, especially those based on convolutional neural networks (CNNs) \cite{csahin2025unlocking}, have significantly improved accuracy through hierarchical feature learning. However, high-performing segmentation networks often come at the cost of increased model complexity and computational overhead \cite{hassan2025comprehensive}.

To bridge the gap between performance and efficiency, lightweight models have gained considerable attention. Approaches like CSRNet~\cite{Xiong2023} and DDRNet~\cite{hong2021deep} employ efficient feature extraction and multi-resolution fusion strategies to balance accuracy and speed. Nonetheless, two fundamental challenges persist: (1) maintaining high accuracy and (2) achieving fast inference for real-time deployment.

To address these issues, lightweight models such as STDC~\cite{stdcnet} and PIDNet~\cite{pidnet} introduce architectural innovations like bidirectional paths and progressive fusion modules, reducing latency while preserving fine-grained details. Similarly, LETNet~\cite{letnet} integrates CNNs and Transformers within a lightweight framework to effectively capture both local and global features. However, transformer-based methods may still encounter limitations in inference cost and memory efficiency on embedded platforms.

Other notable lightweight networks have also emerged. The Parallel Complement Network (PCNet)~\cite{pcnet} introduces a Parallel Complement layer to extract diverse features with large receptive fields using fewer parameters. The Efficient Pyramid Representation Network (EPRNet)~\cite{eprnet} leverages a Multi-scale Processing Unit (MPU) and residual learning to efficiently encode multi-scale features. The Context Aggregation Network~\cite{Yang2021} utilizes a dual-branch design to combine local detail and global context with minimal computational overhead. These developments reflect a growing trend toward designing compact and efficient models tailored for real-time semantic segmentation.

Beyond efficiency, robustness and safety are increasingly critical in real-world, safety-sensitive applications like autonomous driving. Adversarial attacks which are small, imperceptible perturbations designed to deceive neural networks, pose significant threats to the reliability of segmentation models. To mitigate this risk, Edge-Gated Refinement Network (EGRNet) incorporates a lightweight adversarial detection module that analyzes feature-level distributions to identify potential attacks, enhancing overall system safety.

In this paper, we propose EGRNet, a novel lightweight and robust semantic segmentation model that integrates edge-aware refinement and adversarial input detection, specifically tailored for real-time urban scene understanding in autonomous driving. The main contributions of this work are as follows:

\begin{itemize}
    \item We introduce a novel Edge-Gated Refinement (EGR) module that enhances boundary precision by adaptively fusing original and refined features through a learnable gating mechanism focused on edge-sensitive regions.
    
    \item We present EGRNet, a compact and efficient semantic segmentation architecture that achieves state-of-the-art mIoU performance with only \textbf{0.46M} parameters.
    
    \item We incorporate a lightweight adversarial attack detection strategy within the segmentation pipeline, enabling real-time identification of adversarial inputs without impacting overall performance.
\end{itemize}

The remainder of this paper is organized as follows: Section~\ref{sec:related} reviews related work in lightweight semantic segmentation, edge refinement, and adversarial robustness. Section~\ref{sec:method} details the architecture of EGRNet and its core components. Section~\ref{sec:experiments} presents the experimental setup, datasets, and quantitative results, including comparisons with existing methods. Finally, Section~\ref{sec:conclusion} summarizes our findings and discusses potential directions for future research.

\section{Related Work}\label{sec:related}

In the context of autonomous driving, semantic segmentation must operate under real-time constraints while maintaining low computational overhead. Recent efforts have focused on designing lightweight and context-aware architectures to meet these requirements. EGRNet integrates several of these advancements, including depthwise separable convolutions for computational efficiency, squeeze-and-excitation attention for adaptive feature recalibration, dilated residual blocks for multi-scale context extraction, and edge-aware refinement modules for enhanced boundary precision. Furthermore, given the vulnerability of lightweight models to adversarial perturbations, ensuring robustness has become a critical consideration in the design of segmentation networks for safety-critical applications.

\subsection{Lightweight Architectures for Semantic Segmentation}

With the increasing demand for efficient perception in autonomous vehicles, lightweight networks are of interest in recent research. In this context, Shi et al.~\cite{shi2024lightweight} proposed a partial channel transformation strategy that can enhance context representation while maintaining inference speed. By selectively transforming only part of the input feature map, it enables reduced redundancy and real-time performance on embedded systems. FalconNet~\cite{jang2023falcon} also adopts grouped depthwise convolutions to the road scene, which improves accuracy with low latency and is consistent with this direction. Similarly, Junayed et al.~\cite{junayed2022pds} presented PDS-Net, based on the pointwise and depthwise separable layers initially utilized in image detection. On the other hand, they argue that its modular design could be used for segmentation in dynamic environments. Collectively, these studies confirm the direction of relevant architectures towards compact but high-performance versions capable of operating in real-time on autonomous perception tasks.

\subsection{Depthwise Separable Convolutions (DwSConvs)}

Depthwise Separable Convolutions (DwSConvs) are widely used as they are known to substantially reduce the number of parameters and computation with negligible impact on performance. As endorsed by this trend of efficiency, Jang et al.~\cite{jang2023falcon} presented FALCON and showed that DwSConvs can sustain accuracy in real-time applications. This concept was similarly extended by Dbouk and Shanbhag~\cite{dbouk2021generalized}, who introduced generalized DwSConvs and demonstrated improved adversarial robustness as well as model expressiveness?a point further emphasized by Nesti et al.~\cite{nesti2022robustness} in the context of segmentation tasks. Hasan and Dey~\cite{hasan2024depthwise} proposed the idea of deep residual connections augmenting DwSConvs since they believed residual pathways enhance feature reuse while keeping the computation cost low. This is echoed by Dai et al.~\cite{dai2023multi}, who extended DSC with a multi-scale approach based on dilation and demonstrated that concatenating the dilation with DSC allows contextual awareness without sacrificing speed. 

\subsection{Edge-Aware Refinement Techniques}

Precise boundary detection remains a challenge in lightweight segmentation. To tackle this, EGRNet employs a residual gated edge refinement module that borrows from current edge-aware advancements. For instance, Zhang et al.~\cite{zhang2022edge} proposed a gated convolution U-Net variant designed to selectively enhance the boundary features, used for medical imaging, but is adaptable to the urban edge semantics. Dong et al.~\cite{dong2023egfnet} introduced EGFNet, wherein their network fuses RGB and thermal images together using edge-aware guidance to perform close to perfect segmentation even near occlusion. Li et al.~\cite{li2023edge} leveraged edge-aware message passing for fine-grained prediction in forgery detection, demonstrating the versatility of such techniques. Further, Liu et al.~\cite{liu2023edge} considered edge-aware graph autoencoders under imbalanced training for underrepresented road features such as curbs and lane marks. 

\subsection{Attention Mechanisms for Feature Selection}

Channel-wise attention is very important for capturing salient information without expanding network depth in lightweight architectures. With this objective, EGRNet is built upon the "Squeeze-and-Excitation" (SE) blocks to reweight feature maps efficiently~\cite{hu2024aspp}. This is supported by Pang et al.~\cite{pang2021tumor}, who showed that combining channel and spatial attention enhances segmentation accuracy for small-scale structures. Zhang et al.~\cite{zhang2023selective} also argued that Selective Kernel Networks dynamically adjust receptive fields in response to attention cues, improving robustness under various input scales. Additionally, Lv et al.~\cite{lv2022attention} provided a comprehensive overview of attention mechanisms in diagnostic systems, demonstrating their role in increased explainability and spatial localization?important in self-driving applications.

\subsection{Multi-Scale Context Extraction}

Context aggregation is crucial in understanding road scenes with diverse object sizes and scales. Partial channel transformation was used by Shi et al.~\cite{shi2024lightweight} to preserve multi-scale context implicitly. However, most works make use of dilated convolutions or ASPP modules to explicitly extract receptive fields. To maintain global and fine features, Hu et al.~\cite{hu2024aspp} introduced ASPP+-LANet, a hybrid module that combines ASPP with local attention. In low-textural environments such as urban backgrounds, dilated CNNs are shown effective in Liu et al.~\cite{liu2022multi}, especially in spacecraft image segmentation. Following this, Kim et al.~\cite{kim2021multi} optimized this strategy by grouping dilated modules, reducing memory usage while maintaining multi-scale awareness. EGRNet uses dilated residual blocks to ensure contextual consistency between scene elements, building on these prior ideas.

\subsection{Robustness and Adversarial Attacks}

Lightweight models improve speed but often sacrifice robustness, making adversarial conditions a major concern. Gu et al.~\cite{gu2022segpgd} proposed SegPGD, an attack tailored for segmentation networks, showing that standard defenses are weak against fast, inexpensive perturbations. Apostolidis and Papakostas~\cite{apostolidis2021survey} highlighted that even in resource-constrained medical imaging scenarios, robustness must be thoroughly evaluated. On this basis, Nesti et al.~\cite{nesti2022robustness} demonstrated real-world patch attacks against autonomous systems, showing that lightweight networks are particularly vulnerable. Croce et al.~\cite{croce2023robust} responded by proposing training strategies that embed adversarial robustness directly into segmentation pipelines, setting the stage for future work on making EGRNet more resilient.

\section{Proposed Methodology}\label{sec:method}

\begin{figure}
    \centering
    \includegraphics[width=\linewidth]{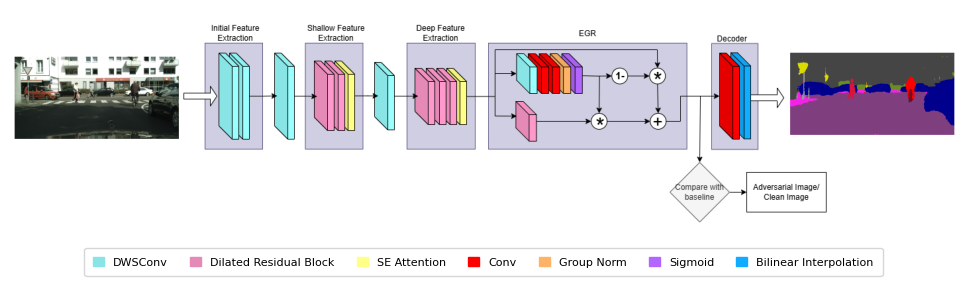}
    \caption{Block diagram of the proposed EGRNet architecture.}
    \label{fig:methodology}
\end{figure}

To achieve a better trade-off between accuracy and efficiency in real-time semantic segmentation, we design a lightweight yet powerful network, EGRNet. The proposed model integrates multiscale context extraction, attention mechanisms, and a custom edge refinement block under a compact encoder-style framework. The overall architecture is illustrated in the block diagram (see Fig.~\ref{fig:methodology}). The detailed layer-wise architecture of EGRNet is summarized in Table~\ref{tab:egrnet}.

\begin{table}[h]
\centering
\caption{Layer-wise architecture of EGRNet. In = Input channels, Out = Output channels, DwSConv = Depthwise Separable Convolution, Dil = Dilation rate, Str = Stride}
\label{tab:egrnet}
\begin{tabular}{@{}lllllll@{}}
\toprule
\textbf{Stage} & \textbf{In} & \textbf{Layer} & \textbf{Out} & \textbf{Dil} & \textbf{Str} & \textbf{Size (H$\times$W)} \\
\midrule

\multirow{2}{*}{Init} 
& 3  & DwSConv              & 48  & 1  & 2 & 256$\times$512 \\
& 48 & DwSConv              & 48  & 1  & 1 & 256$\times$512 \\

Down1 
& 48 & DwSConv              & 96  & 1  & 2 & 128$\times$256 \\

\multirow{3}{*}{Shallow Context}
& 96 & DilatedResBlock      & 96  & 2  & 1 & 128$\times$256 \\
& 96 & DilatedResBlock      & 96  & 4  & 1 & 128$\times$256 \\
& 96 & SE Attention         & 96  & -- & 1 & 128$\times$256 \\

Down2 
& 96 & DwSConv              & 192 & 1  & 2 & 64$\times$128 \\

\multirow{4}{*}{Deep Context}
& 192 & DilatedResBlock     & 192 & 4  & 1 & 64$\times$128 \\
& 192 & DilatedResBlock     & 192 & 8  & 1 & 64$\times$128 \\
& 192 & DilatedResBlock     & 192 & 16 & 1 & 64$\times$128 \\
& 192 & SE Attention        & 192 & -- & 1 & 64$\times$128 \\

EGR 
& 192 & EGR Module          & 192 & 1  & 1 & 64$\times$128 \\

Final 
& 192 & Conv 1$\times$1     & 19  & 1  & 1 & 64$\times$128 \\

Up 
& 19  & Interpolate (bilin.) & 19  & -- & -- & \textbf{512$\times$1024} \\

\bottomrule
\end{tabular}
\end{table}

\subsection{Initial Feature Extraction}

To begin, the input image is passed through an initial feature extraction block composed of two depthwise separable convolutions. The first convolution operates with a stride of 2, effectively reducing the spatial resolution while increasing the number of channels from 3 to 48:
\begin{equation}
    \mathbf{F}_1 = \text{DSConv}_{3 \rightarrow 48}^{3 \times 3, s=2}(\mathbf{I}),
    \label{eq:init1}
\end{equation}
followed by a second depthwise separable convolution that refines the low-level features:
\begin{equation}
    \mathbf{F}_2 = \text{DSConv}_{48 \rightarrow 48}^{3 \times 3}(\mathbf{F}_1).
    \label{eq:init2}
\end{equation}
Here, $\mathbf{I}$ denotes the input image, while $\mathbf{F}_1$ and $\mathbf{F}_2$ are intermediate feature maps extracted at this early stage of the network.

\subsection{Multi-Scale Context Encoding}
To extract multiscale contextual information while preserving efficiency, the network employs a sequence of dilated residual blocks (DRBs) in conjunction with Squeeze-and-Excitation (SE) attention mechanisms. The feature map $\mathbf{F}_2$ is first downsampled using a depthwise separable convolution with stride 2, increasing the number of channels from 48 to 96. Mathematically, 
\begin{equation}
    \mathbf{F}_3 = \text{DSConv}_{48 \rightarrow 96}^{3 \times 3,\ s=2,\ p=1}(\mathbf{F}_2),
    \label{eq:context1}
\end{equation}

This is followed by a context enhancement block comprising two dilated residual blocks with dilation rates $r=2$ and $r=4$, respectively, and an SE attention module applied to the resulting features as,
\begin{equation}
    \mathbf{F}_4 = \text{SE}\left( \text{DRB}^{r=4}_{96} \left( \text{DRB}^{r=2}_{96}(\mathbf{F}_3) \right) \right),
    \label{eq:context2}
\end{equation}
where $\text{DRB}^{r}_{c}$ denotes a dilated residual block with dilation rate $r$ and $c$ channels, and SE denotes a squeeze-and-excitation attention mechanism.

Subsequently, the network further downsamples the feature map to deepen the representation, this time increasing the channel dimension from 96 to 192, i.e., 
\begin{equation}
    \mathbf{F}_5 = \text{DSConv}_{96 \rightarrow 192}^{3 \times 3,\ s=2,\ p=1}(\mathbf{F}_4),
    \label{eq:context3}
\end{equation}

The resulting high-level features $\mathbf{F}_5$ are refined through a deeper context block composed of three sequential DRBs with increasing dilation rates ($r=4$, $r=8$, and $r=16$), followed by another SE module.
\begin{equation}
    \mathbf{F}_6 = \text{SE}\left( \text{DRB}^{r=16}_{192} \left( \text{DRB}^{r=8}_{192} \left( \text{DRB}^{r=4}_{192}(\mathbf{F}_5) \right) \right) \right).
    \label{eq:context4}
\end{equation}

These stages enable the network to efficiently model long-range dependencies without incurring excessive computational cost.

\subsection{Edge-Gated Refinement Module (EGR)}
To enhance spatial details and preserve object boundaries, especially after successive downsampling, we introduce the EGR module. This block integrates a gated attention mechanism that adaptively combines original and refined features. First, a gating map $\mathbf{G}$ is computed by applying a series of convolutions followed by group normalization and a sigmoid activation, as shown in~\ref{eq:gating}.
\begin{equation}
    \mathbf{G} = \sigma(\text{GN}(\text{Conv}_{1 \times 1}(\text{GatedConv}_{3 \times 3}(\text{Conv}_{3 \times 3}(\mathbf{F}_6))))),
    \label{eq:gating}
\end{equation}

Simultaneously, a refined representation $\mathbf{R}$ is generated via a depthwise separable convolution as,
\begin{equation}
    \mathbf{R} = \text{DSConv}_{192 \rightarrow 192}^{3 \times 3}(\mathbf{F}_6),
    \label{eq:refine}
\end{equation}

\begin{figure}[t]
    \centering
    \includegraphics[width=\linewidth]{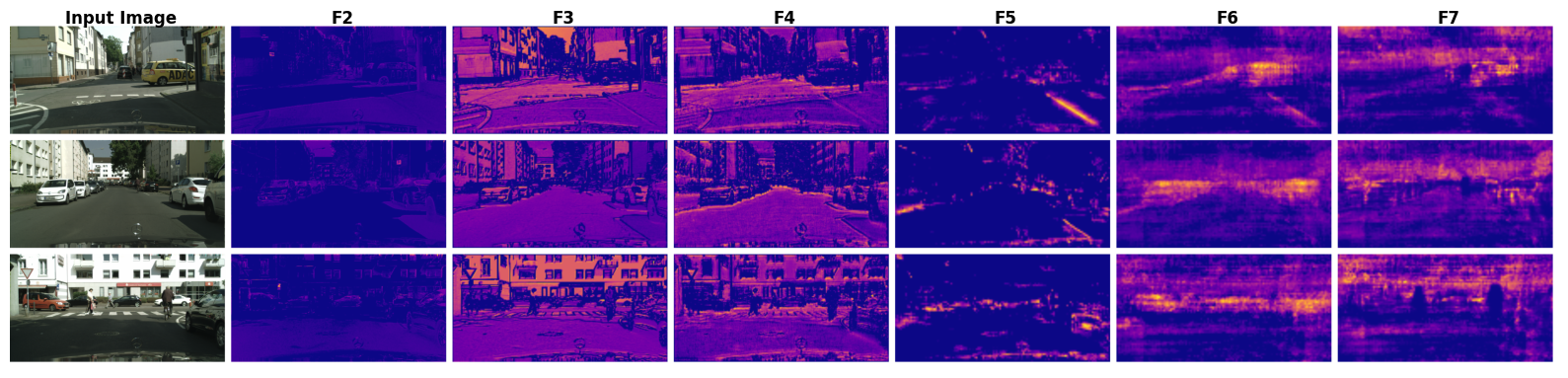}
    \caption{Visualization of feature maps at key layers in the EGRNet model for a sample input image. The first image in each row is the original input, followed by the feature maps at various stages.} 
    \label{fig:feature_maps}
\end{figure}

The final output of the EGR module, $\mathbf{F}_7$, is obtained by blending the refined and original features using the learned gating map, as formulated in \ref{eq:blend}.
\begin{equation}
    \mathbf{F}_7 = \mathbf{G} \odot \mathbf{R} + (1 - \mathbf{G}) \odot \mathbf{F}_6,
    \label{eq:blend}
\end{equation}
where $\sigma(\cdot)$ denotes the sigmoid function, $\odot$ represents element-wise multiplication, and $\text{GN}$ indicates group normalization. Feature maps at key layers of EGRNet are visualized in Fig.~\ref{fig:feature_maps}. The first image in each row is the original input, followed by the feature maps at various stages. These visualizations correspond to the intermediate representations defined in \ref{eq:init2}--\ref{eq:blend}.

\subsection{Final Prediction Layer}
A $1 \times 1$ convolution is applied to the refined feature map to project it onto the target number of semantic classes as,
\begin{equation}
    \mathbf{P} = \text{Conv}_{1 \times 1}^{192 \rightarrow C}(\mathbf{F}_7),
    \label{eq:logits}
\end{equation}
where $C$ denotes the number of output classes. To recover the original input resolution, the resulting logits $\mathbf{P}$ are upsampled using bilinear interpolation. Mathematically,
\begin{equation}
    \mathbf{P}_{\text{final}} = \text{Upsample}_{\text{bilinear}}(\mathbf{P}, \text{size}=(512, 1024)).
    \label{eq:upsample}
\end{equation}

\subsection{Detection of adversarial inputs}

The activation output of a specified intermediate layer is crucial for understanding the model's internal representation of input data. To achieve this, we modify the trained EGRNet to allow for the extraction of activations from the EGR module. A forward hook function is registered to intercept the output of the target layer during the forward pass. This hook function stores the activation for further analysis. Once the hook is registered, the model processes the input image, and the activation from the EGR module is captured and returned for use in the anomaly detection process.

\begin{figure}
    \centering
    \includegraphics[width=\linewidth]{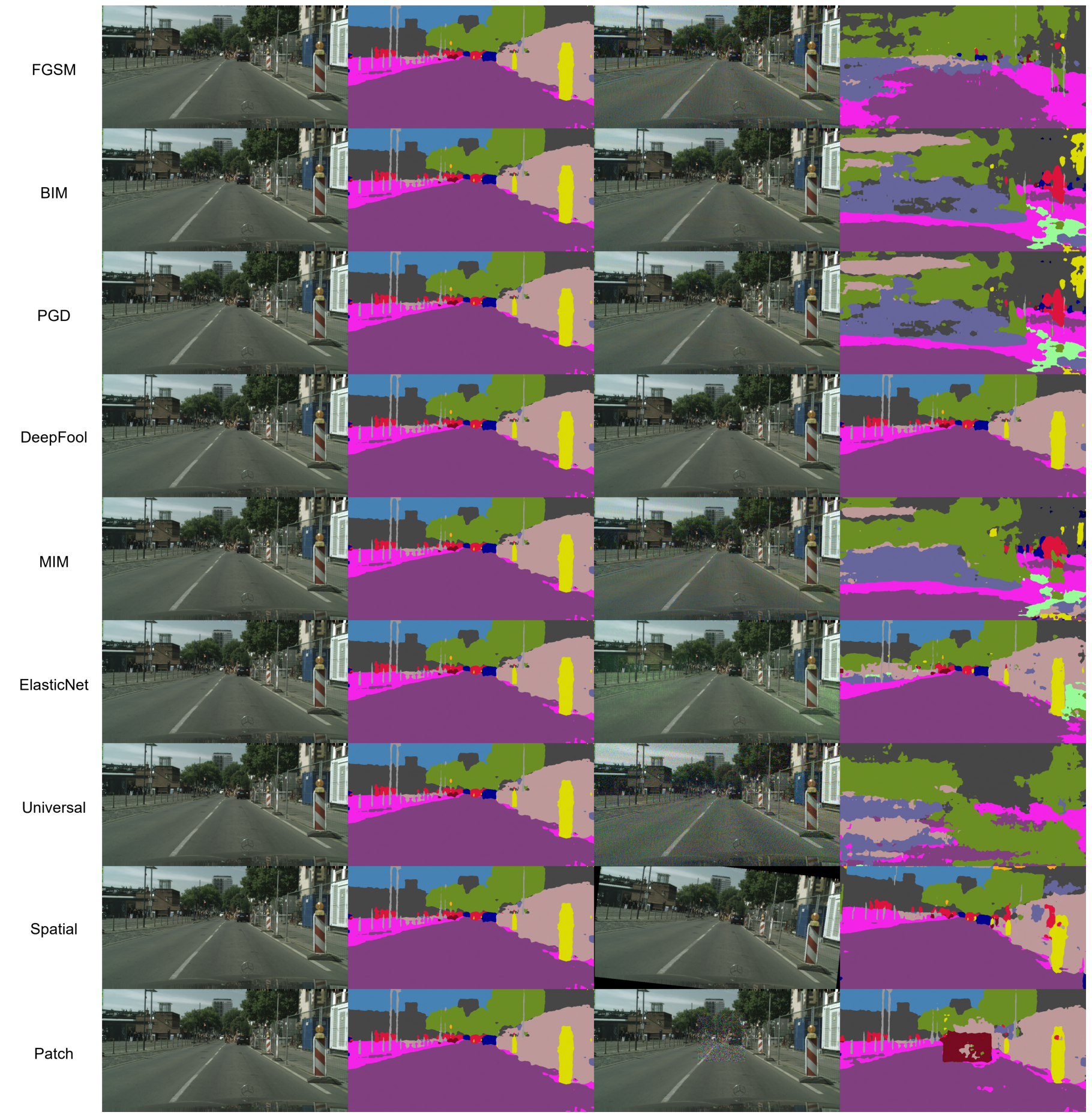}
    \caption{Visualization of EGRNet predictions under different adversarial attacks. Each row shows the original input and its prediction (left), followed by the adversarial input and its corresponding prediction (right) for a specific attack. This highlights the effect of various adversarial perturbations on model performance.}
    \label{fig:adversarial_attacks}
\end{figure}

Then we analyze the activations extracted in the previous step. An anomaly is defined as a significant deviation from the norm, which is quantified using the L2 norm of the activations. The L2 norm provides a measure of the overall intensity of the neural responses in the intermediate layer. To detect anomalies, we compute the L2 norm of the extracted activation and compare it to a reference distribution characterized by its mean and standard deviation. The reference mean and standard deviation are computed from the activations of a set of normal (non-adversarial) inputs, which were obtained from the clean training set, prior to performing anomaly detection. The difference between the current activation norm and the reference mean is calculated, and if this difference exceeds a predefined threshold, the input is flagged as anomalous. The threshold is set to 2 times the standard deviation of the reference norm. If the norm difference exceeds this threshold, the image is considered anomalous, potentially indicating the presence of adversarial manipulation. Otherwise, the input is classified as benign. Fig.~\ref{fig:adversarial_attacks} illustrates the effect of different adversarial attacks on the model's performance. Each row shows the original image with its prediction, followed by the adversarially perturbed input and its corresponding prediction. This figure visually demonstrates how adversarial inputs can significantly disrupt the model's predictions, which is crucial for understanding how the anomaly detection system operates. The steps involved in the detection are illustrated in Algorithm \ref{alg:adversarial_detection} which shows that the model is first initialized, and the target layer for activation extraction is identified. A forward hook is registered to capture the activation output during the forward pass. The activation's L2 norm is computed and compared with a reference norm. If the difference between the activation's norm and the reference mean exceeds the threshold (set to 2 times the standard deviation), the input is flagged as anomalous.

\begin{algorithm}[t]
\caption{Adversarial Input Detection using Intermediate Layer Activations}
\label{alg:adversarial_detection}
\begin{algorithmic}[1]
\Require Pretrained model $model$, target layer $target\_layer$, input image $x$, mean activation norm $\mu$, standard deviation $\sigma$, threshold $T$
\Ensure Boolean flag indicating adversarial input

\State Initialize $model$ with $target\_layer$ for activation extraction
\State Register a forward hook to capture activation at $target\_layer$
\State Run forward pass on $x$ to obtain activation $f(x)$
\State Compute $A \Leftarrow \|f(x)\|_2$
\State Compute $D \Leftarrow |A - \mu|$

\If{$D > T \cdot \sigma$}
    \State \Return \texttt{True}
\Else
    \State \Return \texttt{False}
\EndIf

\end{algorithmic}
\end{algorithm}

\section{Experimental Results and Analysis}\label{sec:experiments}

We conduct all experiments on the Cityscapes dataset~\cite{cordts2016cityscapes}, a large-scale benchmark for semantic urban scene understanding. It contains 5,000 finely annotated images of size 1024$\times$2048, distributed across 19 semantic classes such as roads, pedestrians, vehicles, and buildings. To balance performance and efficiency, we resize input images to 512$\times$1024. All models are evaluated on the Cityscapes \textit{validation set}, which serves as our test set to assess generalization performance. We compare four semantic segmentation models: DABNet \cite{Li2019}, LCNet \cite{shi2024lightweight}, LMFFNet\cite{Shi2023}, and our proposed EGRNet. A seen in Table~\ref{tab:class_category_ious}, which reports per-class IoUs, along with class-wise and category-wise mean IoUs, our model achieves the best performance across several challenging classes (e.g., wall, fence, train, motorcycle, truck, bus), indicating superior boundary refinement and contextual understanding.

\begin{sidewaystable}
\caption{Per-class IoU (\%) on Cityscapes validation set. CmIoU = Class mIoU, CatmIoU = Category mIoU.}
\label{tab:class_category_ious}

\tiny
\setlength{\tabcolsep}{3pt}

\begin{tabular*}{\textheight}{@{\extracolsep\fill}lccccccccccccccccccccc}
\toprule
\textbf{Method} & Rd & Sd & Bld & Wl & Fn & Pl & TL & TS & Veg & Ter & Sky & Per & Rid & Car & Trk & Bus & Trn & Mot & Bik & CmIoU & CatmIoU \\
\midrule

DABNet 
& 96.1 & 71.5 & 86.4 & 32.8 & 41.0 & 41.6 & 40.3 & 55.5 & 87.7 & 47.9 & 90.6 & 62.4 & 37.8 & 87.8 & 27.0 & 43.3 & 34.9 & 23.7 & 59.0 & 56.2 & 62.5 \\

LCNet 
& \textbf{96.4} & 73.8 & \textbf{87.8} & 42.5 & 43.1 & \textbf{47.3} & \textbf{44.0} & 56.7 & 89.1 & 55.4 & 91.7 & \textbf{64.9} & 37.0 & 87.8 & 36.8 & 51.4 & 26.1 & 24.2 & \textbf{60.1} & 58.7 & 65.0 \\

LMFFNet 
& \textbf{96.6} & \textbf{75.1} & 88.1 & 43.3 & 41.9 & 48.4 & 44.5 & \textbf{59.8} & \textbf{89.3} & 54.3 & 91.5 & 67.6 & \textbf{43.7} & \textbf{90.3} & 50.0 & 59.1 & 41.3 & 27.3 & 63.6 & 61.9 & 67.0 \\

\textbf{EGRNet} 
& 96.2 & 74.0 & 87.1 & \textbf{49.6} & \textbf{51.7} & 41.4 & 39.5 & 56.2 & 88.4 & \textbf{64.4} & \textbf{92.0} & 62.5 & 38.3 & 89.3 & \textbf{67.4} & \textbf{76.5} & \textbf{69.9} & \textbf{35.9} & 60.0 & \textbf{65.3} & \textbf{68.4} \\

\bottomrule
\end{tabular*}
\end{sidewaystable}

\begin{figure}[htbp]
    \centering
    \includegraphics[width=0.45\linewidth]{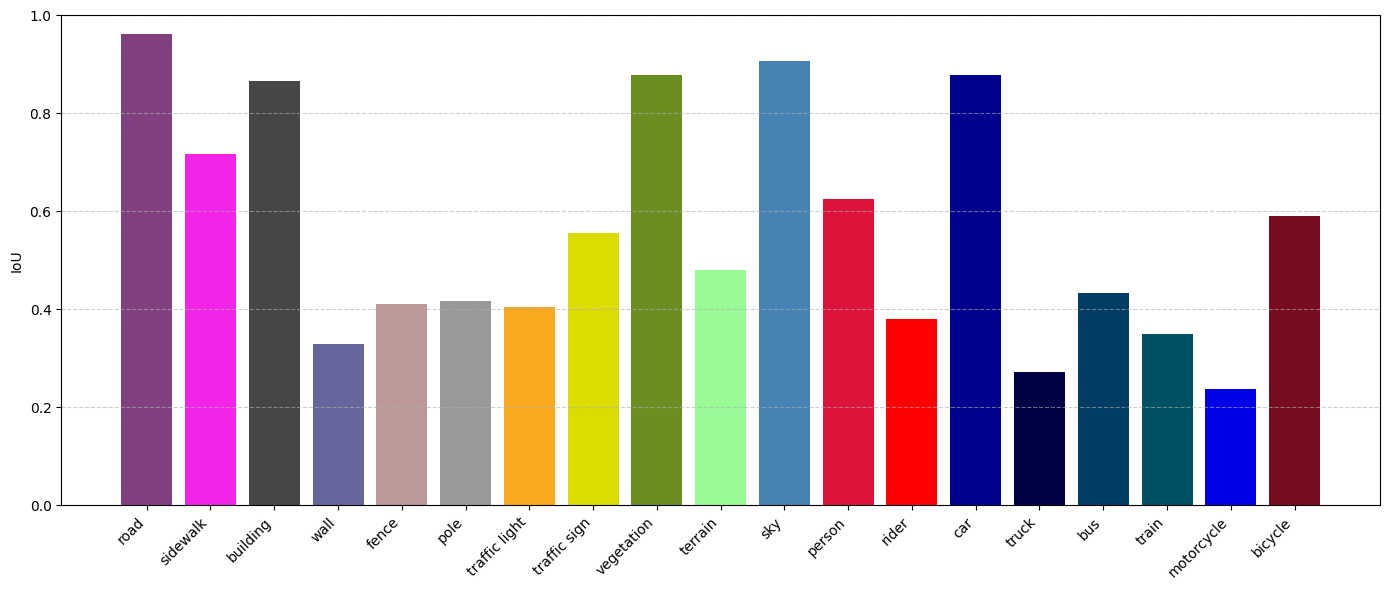}
    \includegraphics[width=0.45\linewidth]{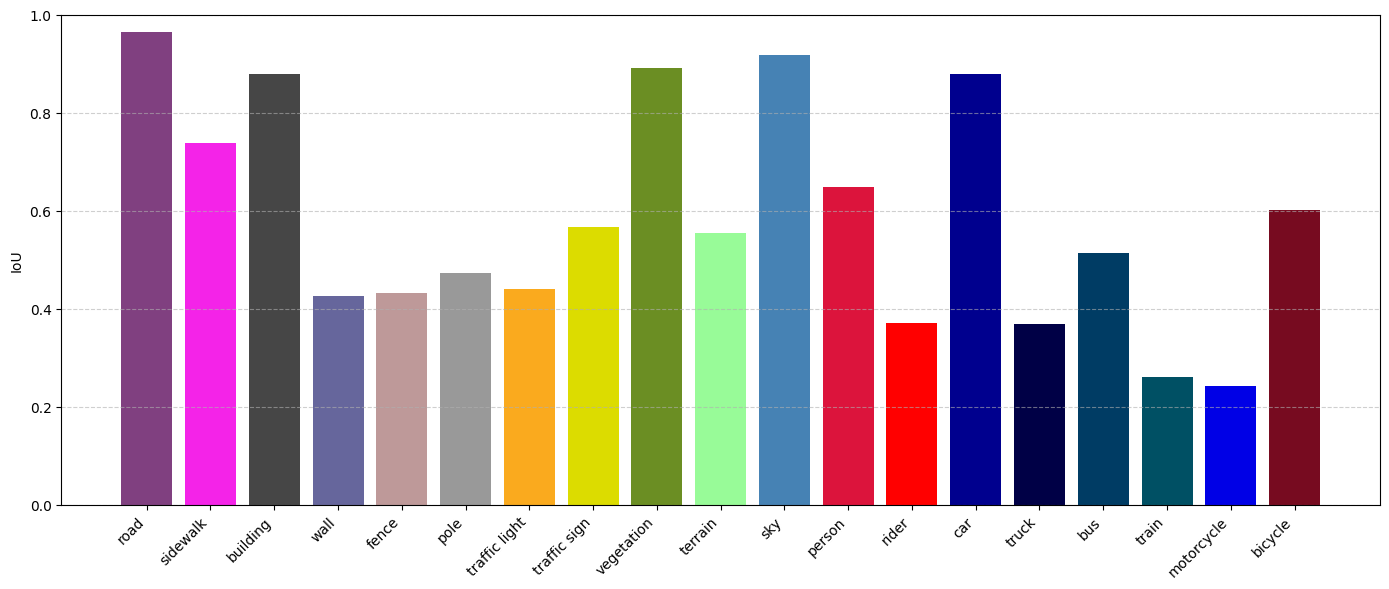} \\
    \includegraphics[width=0.45\linewidth]{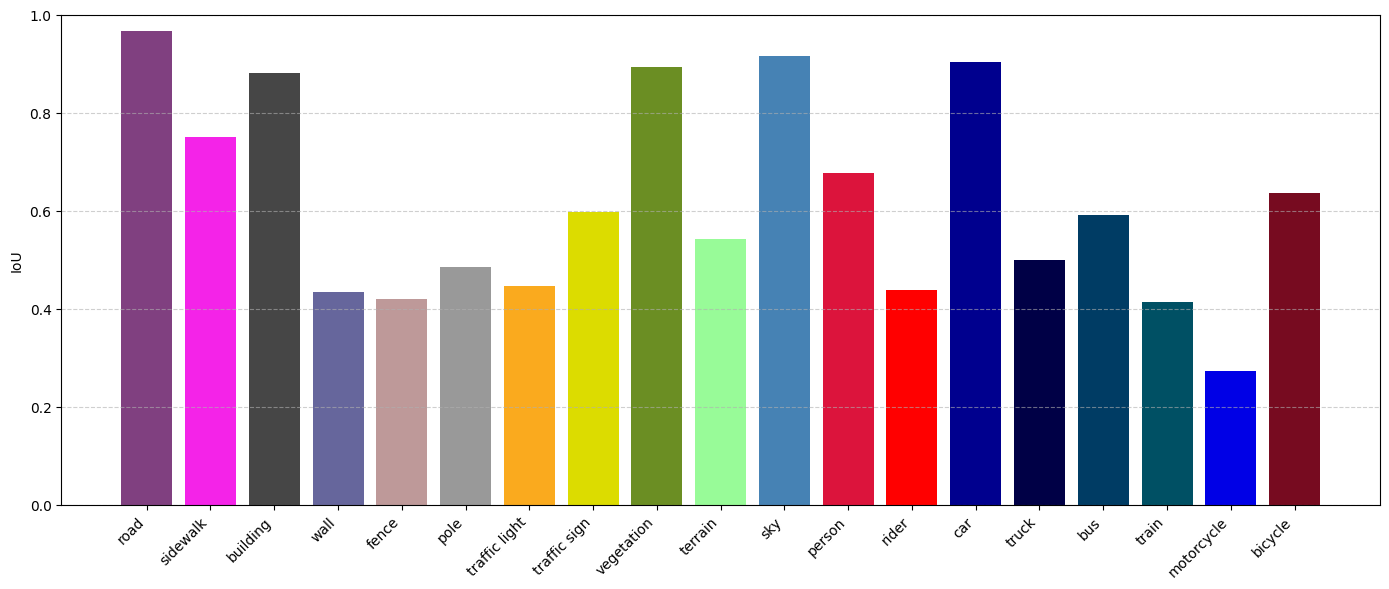}
    \includegraphics[width=0.45\linewidth]{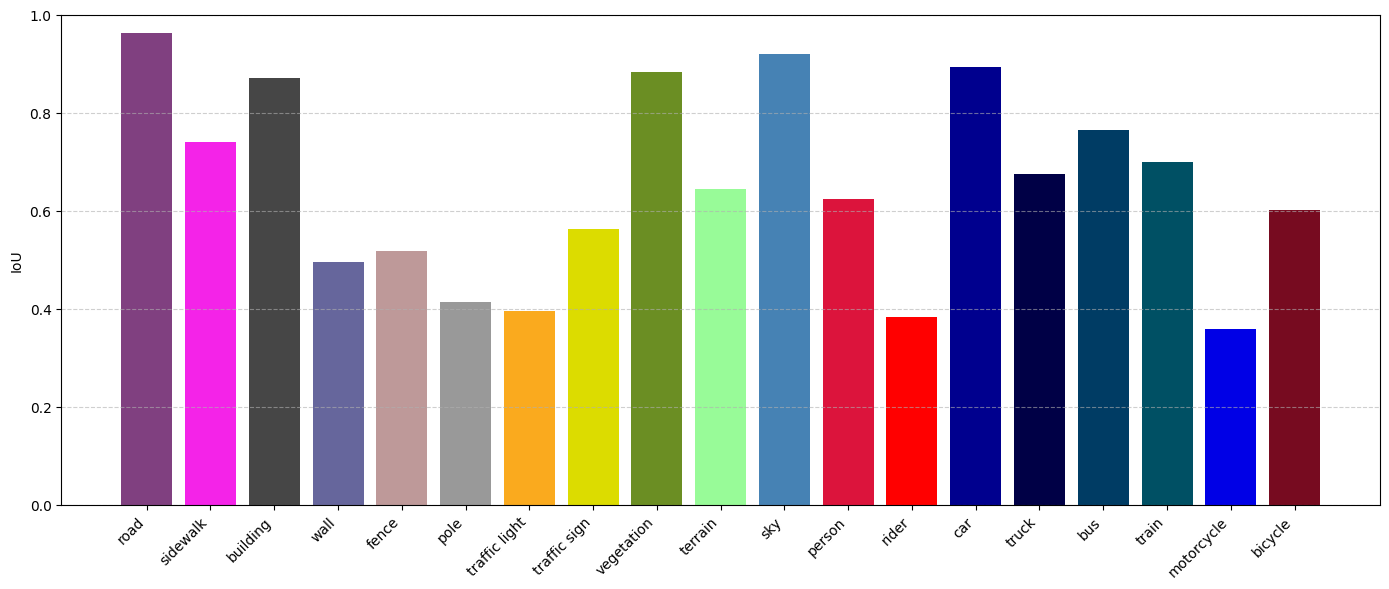}
    \caption{Per-class IoU scores across all models. From top-left to bottom-right: DABNet, LCNet, LMFFNet, and EGRNet (Ours).}
    \label{fig:per_class_ious}
\end{figure}

\begin{figure}[htbp]
    \centering
    \includegraphics[width=0.45\linewidth]{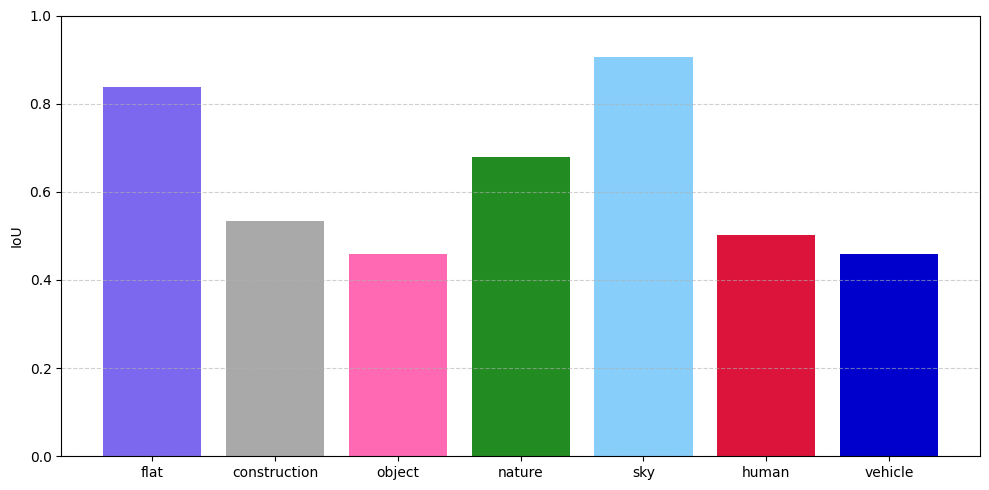}
    \includegraphics[width=0.45\linewidth]{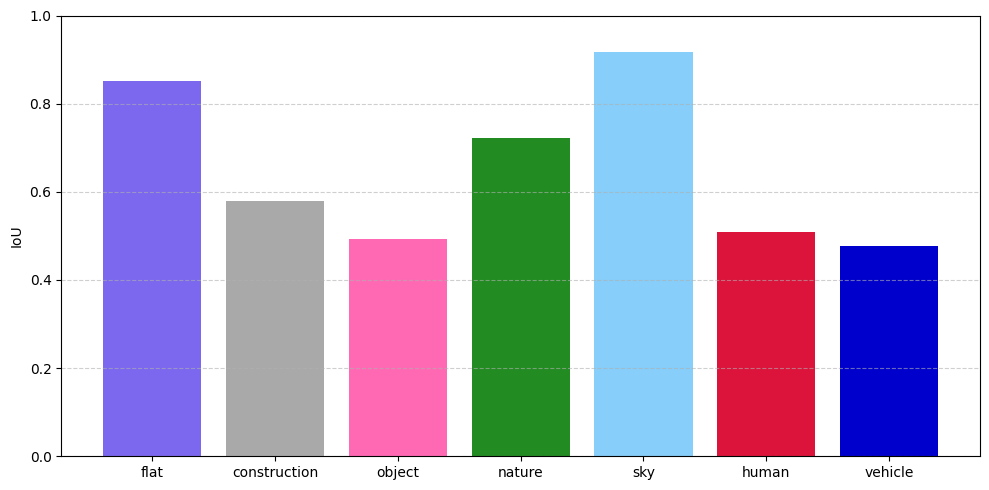} \\
    \includegraphics[width=0.45\linewidth]{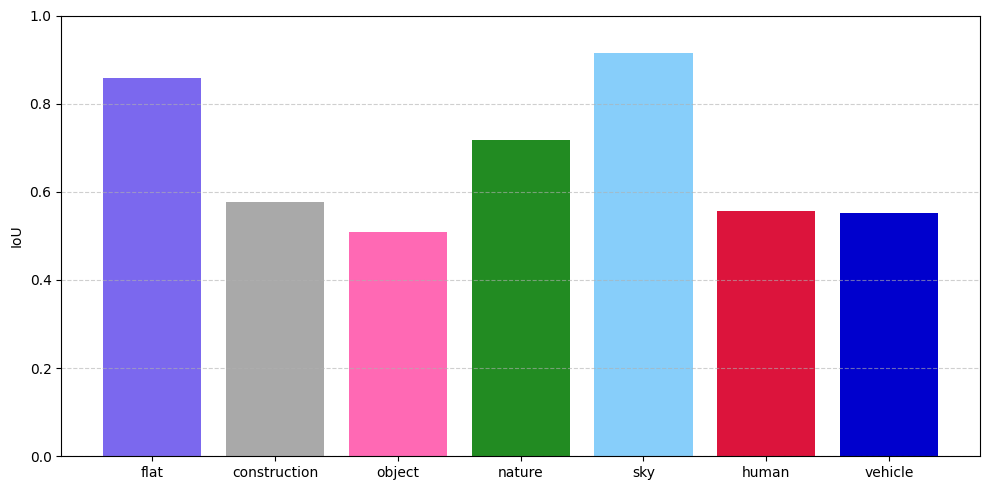}
    \includegraphics[width=0.45\linewidth]{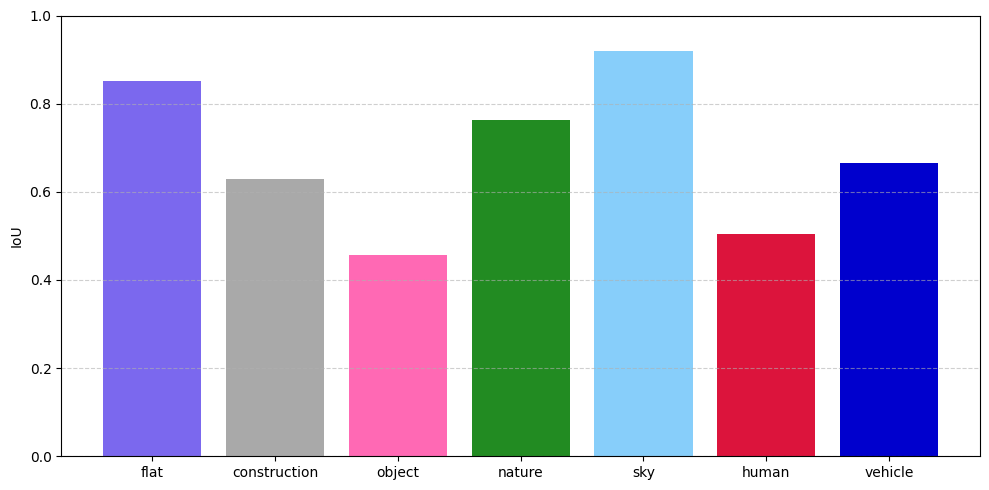}
    \caption{Category-wise IoU scores for all models. From top-left to bottom-right: DABNet, LCNet, LMFFNet, and EGRNet (Ours).}
    \label{fig:category_ious}
\end{figure}

The plots in Fig.~\ref{fig:per_class_ious} illustrate how each model performs across individual semantic classes. Whereas the plots in Fig.~\ref{fig:category_ious} demonstrate the strengths of each model across broader semantic categories. EGRNet demonstrates a balanced performance across both frequent and rare classes, particularly excelling in object boundaries and fine-grained details.

\begin{figure}[htbp]
    \centering
    \includegraphics[width=0.6\linewidth]{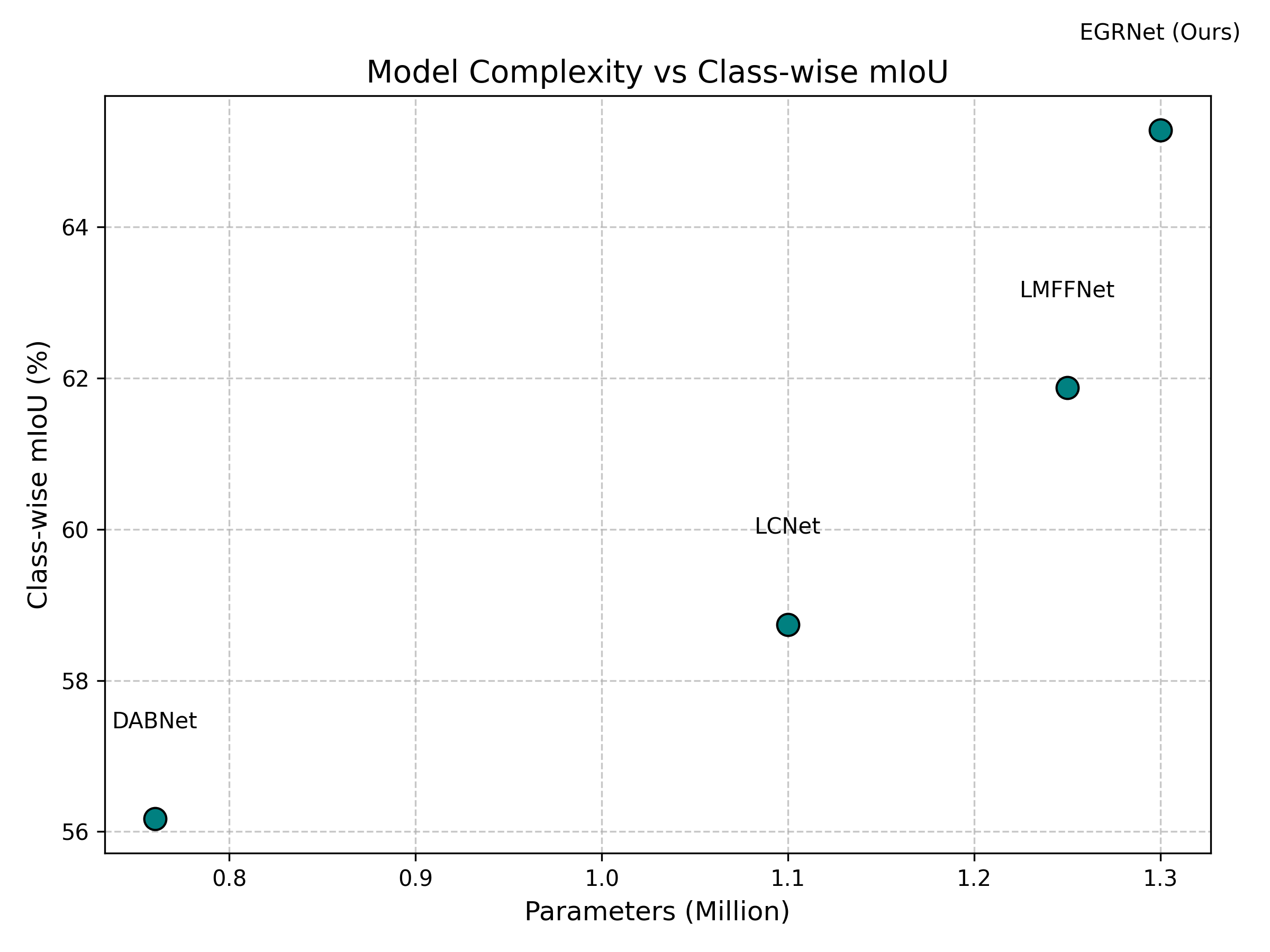}
    \caption{Comparison between number of parameters and mIoU. EGRNet demonstrates strong segmentation performance while remaining lightweight.}
    \label{fig:params_vs_miou}
\end{figure}

\begin{figure}[t]
    \centering
    \includegraphics[width=\linewidth]{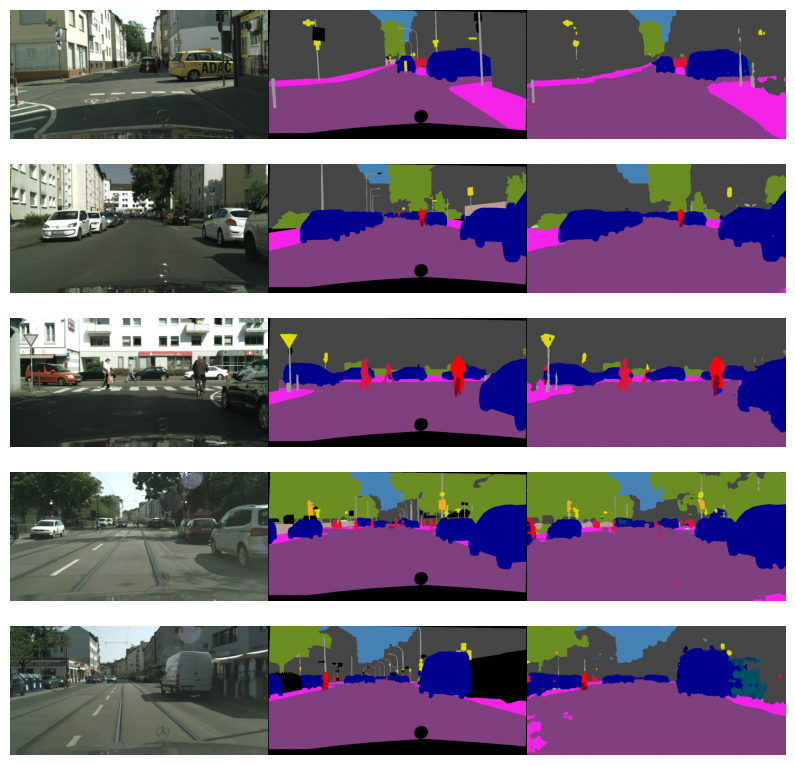}
    \caption{Qualitative segmentation results for selected images.}
    \label{fig:segmentation_results}
\end{figure}

Table~\ref{tab:complexity_miou} shows that EGRNet achieves state-of-the-art accuracy while being the most lightweight among all compared models. This can also be visualized in Fig.~\ref{fig:params_vs_miou}. The qualitative results shown in Fig.~\ref{fig:segmentation_results} confirm that EGRNet consistently produces sharp boundaries and semantically coherent predictions, especially in crowded or cluttered scenes.


\begin{table}
\centering
\caption{Comparison of model complexity and segmentation performance.}
\label{tab:complexity_miou}
\begin{tabular}{@{}lccc@{}}
\toprule
\textbf{Model} & \textbf{Params (M)} & \textbf{MACs (G)} & \textbf{Class-wise mIoU (\%)} \\
\midrule
DABNet & 0.76 & 10.56 & 56.17 \\
LCNet & 0.51 & 7.92 & 58.74 \\
LMFFNet & 1.35 & 16.66 & 61.87 \\
\textbf{EGRNet (Ours)} & \textbf{0.46} & \textbf{5.00} & \textbf{65.28} \\
\bottomrule
\end{tabular}
\end{table}

\begin{figure}
    \centering
    \includegraphics[width=0.7\linewidth]{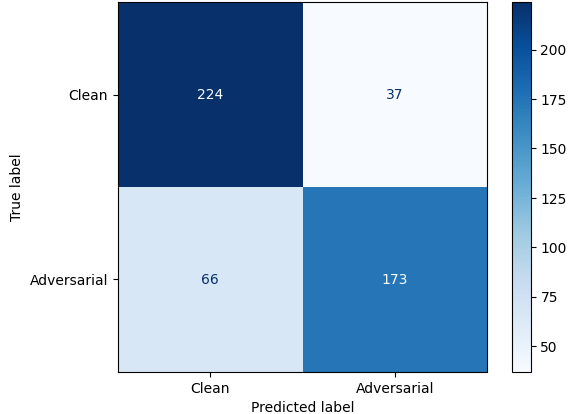}
    \caption{Confusion matrix for adversarial detection. It shows detection accuracy across clean and adversarial inputs.}
    \label{fig:confusion_matrix}
\end{figure}

To evaluate the robustness of EGRNet in detecting adversarial inputs, we assess its performance on adversarial examples generated using nine widely adopted attack methods: Fast Gradient Sign Method (FGSM) \cite{goodfellow2015explaining}, Basic Iterative Method (BIM) \cite{kurakin2017adversarial}, Projected Gradient Descent (PGD) \cite{madry2018towards}, DeepFool \cite{moosavi2016deepfool}, Momentum Iterative Method (MIM) \cite{dong2018boosting}, Elastic-Net Attack \cite{chen2018ead}, Universal Adversarial Perturbation (UAP) \cite{moosavi2017universal}, Spatial Transformation Attack \cite{xiao2018spatial}, and Adversarial Patch Attack \cite{brown2017adversarial}. FGSM is a simple yet effective single-step gradient-based attack that perturbs input data in the direction of the gradient to maximize the model?s loss. BIM extends FGSM by applying it iteratively with small step sizes, thereby producing more powerful perturbations. PGD builds upon BIM by including a projection step to keep perturbations within a defined bound, and is considered one of the strongest first-order attacks. MIM enhances BIM by incorporating momentum, which helps escape poor local optima and improves attack transferability across models. DeepFool is an untargeted attack that iteratively pushes the input across the decision boundary with minimal perturbation. Elastic-net attack introduces elastic-net regularization by combining L1 and L2 penalties to generate sparse yet effective perturbations. UAP generates a universal, input-agnostic perturbation capable of fooling the model across many different inputs. Spatial Transformation Attack modifies spatial properties such as rotation or translation, rather than directly altering pixel values, making it harder to detect using conventional pixel-level defenses. The Adversarial Patch Attack places a visually noticeable patch on the input image that causes misclassification and is especially relevant in physical-world scenarios. By evaluating EGRNet against this diverse set of attacks, we can comprehensively analyze its ability to detect and respond to both subtle and conspicuous adversarial manipulations. These adversarial samples are crafted on the Cityscapes validation set, which is used exclusively for testing and is not seen during training. The objective is to analyze EGRNet's ability to correctly identify and distinguish adversarial inputs from clean ones, thereby evaluating its resilience against real-world adversarial threats. The confusion matrix in Fig.~\ref{fig:confusion_matrix} summarizes the classification outcomes. It captures both true and false classification rates across clean and attacked images.

These results demonstrate the feasibility of integrating lightweight adversarial detectors within segmentation pipelines. Nonetheless, further improvements in threshold tuning and uncertainty modeling could reduce false alarm rates.

\section{Conclusion}\label{sec:conclusion}

In this work, we introduced EGRNet, a lightweight yet robust semantic segmentation network designed specifically for real-time autonomous driving applications. By leveraging depthwise separable convolutions, squeeze-and-excitation attention mechanisms, dilated residual blocks, and a novel edge-aware refinement module, EGRNet achieves high computational efficiency while maintaining precise boundary segmentation. Our extensive experiments on the Cityscapes validation set demonstrate that EGRNet outperforms several state-of-the-art lightweight models, delivering competitive segmentation accuracy with significantly lower computational costs. Additionally, EGRNet demonstrates strong resilience to a wide range of adversarial attacks--including FGSM, BIM, PGD, DeepFool, MIM, ElasticNet, Universal, Spatial, and Patch--successfully detecting adversarial inputs through intermediate activation analysis. This positions EGRNet as a reliable and safe perception model for autonomous driving systems. Qualitative results further highlight its ability to produce sharp and semantically meaningful predictions in complex urban environments. Overall, EGRNet offers a promising solution for efficient and robust semantic segmentation, with future work focused on developing methods to not only detect but also mitigate adversarial perturbations, improve model interpretability, and evaluate real-time deployment in on-board vehicle platforms.

\bibliographystyle{unsrtnat}
\bibliography{references}

\end{document}